\PassOptionsToPackage{table}{xcolor}
\documentclass[10pt,twocolumn,letterpaper]{article}

\usepackage[pagenumbers]{cvpr}
\usepackage{array}
\usepackage{tabularx}
\usepackage{colortbl}
\usepackage{tikz}
\usepackage{float}
\usepackage{makecell}
\usepackage{multirow}
\usepackage{wrapfig}
\usepackage{pifont}

\usetikzlibrary{arrows.meta,positioning,fit,calc}

\newcommand{\cmark}{\textcolor{green!70!black}{\ding{51}}}
\newcommand{\xmark}{\textcolor{red!80!black}{\ding{55}}}

\newcolumntype{L}[1]{>{\raggedright\arraybackslash}p{#1}}
\newcolumntype{C}[1]{>{\centering\arraybackslash}p{#1}}
\newcolumntype{Y}{>{\raggedright\arraybackslash}X}

\colorlet{surfA}{gray!05}
\colorlet{embedA}{RoyalBlue!05}
\colorlet{cycleA}{Orange!8}
\colorlet{oursA}{ForestGreen!10}

\colorlet{high1}{red!25}
\colorlet{high2}{yellow!25}

\colorlet{surfB}{gray!35}
\colorlet{embedB}{RoyalBlue!45}
\colorlet{cycleB}{Orange!55}
\colorlet{oursB}{ForestGreen!60}

\definecolor{cvprblue}{rgb}{0.21,0.49,0.74}
\usepackage[pagebackref,breaklinks,colorlinks,allcolors=cvprblue]{hyperref}

\def\paperID{*****}
\def\confName{CVPR}
\def\confYear{2026}

\title{POVQA: Preference-Optimized Video Question Answering with Rationales for Data Efficiency}

\author{Ashim  Dahal \quad Ankit Ghimire \quad Saydul Akbar Murad \quad Nick Rahimi\\
University of Southern Mississippi\\
Hattiesburg, Mississippi, USA\\
{\tt\small \{ashim.dahal,ankit.ghimire,saydulakbar.murad,nick.rahimi\}@usm.edu}}

\begin{document}
\twocolumn[{%
\renewcommand\twocolumn[1][]{#1}%
\maketitle
\begin{center}
    \centering
    \captionsetup{type=figure}
    \includegraphics[width=\textwidth]{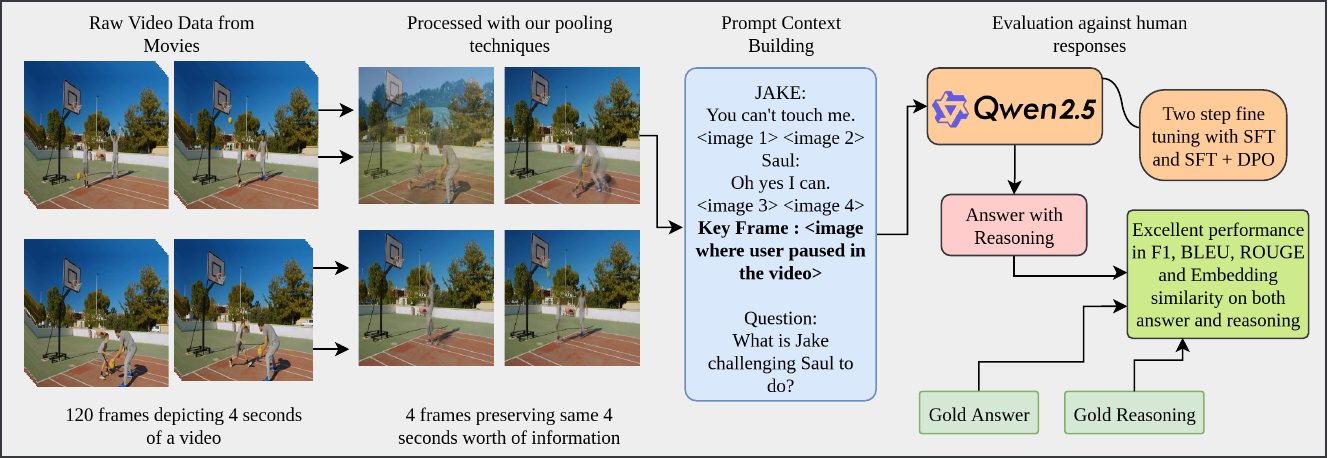}
    \captionof{figure}{Overview of POVQA inference on the ReasonVQA evaluation split.}
    \label{fig:abstract_figure}
\end{center}%
}]

\begin{abstract}
Long-video multimodal question answering requires structured reasoning over visual evidence and dialogue, but Large Vision-Language Models (LVLMs) are constrained by context-window and compute limits. We propose POVQA, which compresses each second into a temporally pooled image (1 fps pooled images) to maintain dense temporal coverage under a fixed token budget. We then train Qwen2.5-VL-7B with supervised fine-tuning (SFT) on rationale+answer targets, and optionally apply Direct Preference Optimization (DPO) for preference alignment. We introduce ReasonVQA as a pilot diagnostic dataset with 12 movies and 239 human-annotated QA+rationale triplets for controlled analysis of long-context multimodal reasoning under compression. On ReasonVQA, SFT improves the best pooled-only baseline from 0.212 to 0.550 F1, showing that pooled evidence plus rationale supervision provides the main performance gains in this setting. In zero-shot transfer, POVQA also reaches 64.7\% on TVQA after SFT+DPO. These results are preliminary: ReasonVQA is small, pooling can lose fine-grained temporal order, and DPO effects are not uniformly positive across settings. Code, dataset, and additional qualitative evaluations are available at \href{https://povqa.github.io}{https://povqa.github.io}.
\end{abstract}

\section{Introduction}
Video question answering (VQA) over movies and TV episodes requires combining dialogue, scene context, and temporally distributed visual evidence~\cite{lei2018tvqa,tapaswi2016movieqa}. This is a multimodal algorithmic reasoning setting: the model must integrate language and vision across time, then produce a structured answer grounded in evidence. In practice, long-video VQA remains difficult for modern Large Vision-Language Models (LVLMs): even large context windows represent only a short clip at native frame rates, while compute and memory grow quickly with video length~\cite{alayrac2022flamingo,chen2023pali,llava,Maaz2023VideoChatGPT,ren2024timechat,song2024moviechat,yang2025qwen2}.

A multi-minute clip contains thousands of frames. Feeding all frames is often infeasible, but aggressive frame dropping can remove key evidence for causal or temporal questions. This creates a core tradeoff between temporal coverage and token budget.

We study a simple long-context pipeline, POVQA, that compresses video into 1 fps pooled images. Instead of selecting a few isolated frames, each pooled image summarizes a full second using blend-blur and weighted averaging variants. We then interleave pooled images with subtitle text and prompt an LVLM to generate both a rationale and a final answer, following the broader multimodal interleaving direction in prior works~\cite{Jiang2024MANTISIM,li2024llava,Tian2024MMInterleavedII}.

Pooling acts as a temporal evidence accumulator: each token encodes changes across many raw frames while keeping a fixed input length. This can preserve broad motion and scene transitions under a constrained budget, though it may fail on order-sensitive micro-actions. In this sense, POVQA targets test-time reasoning efficiency rather than architecture scaling.

We train with supervised fine-tuning (SFT) on rationale+answer outputs, then optionally apply Direct Preference Optimization (DPO). Rationale supervision encourages explicit evidence use, while DPO can align output style toward preferred responses in some regimes.

We use Qwen2.5-VL-7B for open weights, a strong vision-language backbone, and reproducible tooling. We do not claim this is the most efficient model; our focus is the input-compression and alignment recipe, which can later be tested on smaller backbones.

\noindent\textbf{Contributions.}
\begin{itemize}
    \item A long-context multimodal reasoning pipeline that compresses long videos into a fixed-budget sequence of temporally pooled evidence images at 1 fps.
    \item A rationale-supervised SFT setup and a DPO variant, with analysis of where DPO helps or hurts in a small-data setting.
    \item ReasonVQA: a small pilot diagnostic dataset (12 movies, 239 QA+rationale triplets) for controlled evaluation of long-video reasoning under compression, including explicit failure-mode analysis.
\end{itemize}

\noindent\textbf{Scope and limitations.} This work is intentionally scoped as a pilot study, not a large benchmark release. ReasonVQA is small and movie-centric, pooling can miss fine-grained temporal order, and DPO can be unstable under limited preference data. We therefore position results as controlled evidence for this design choice rather than broad claims of generalization.

\begin{figure*}[t]
    \centering
    \includegraphics[width=\linewidth]{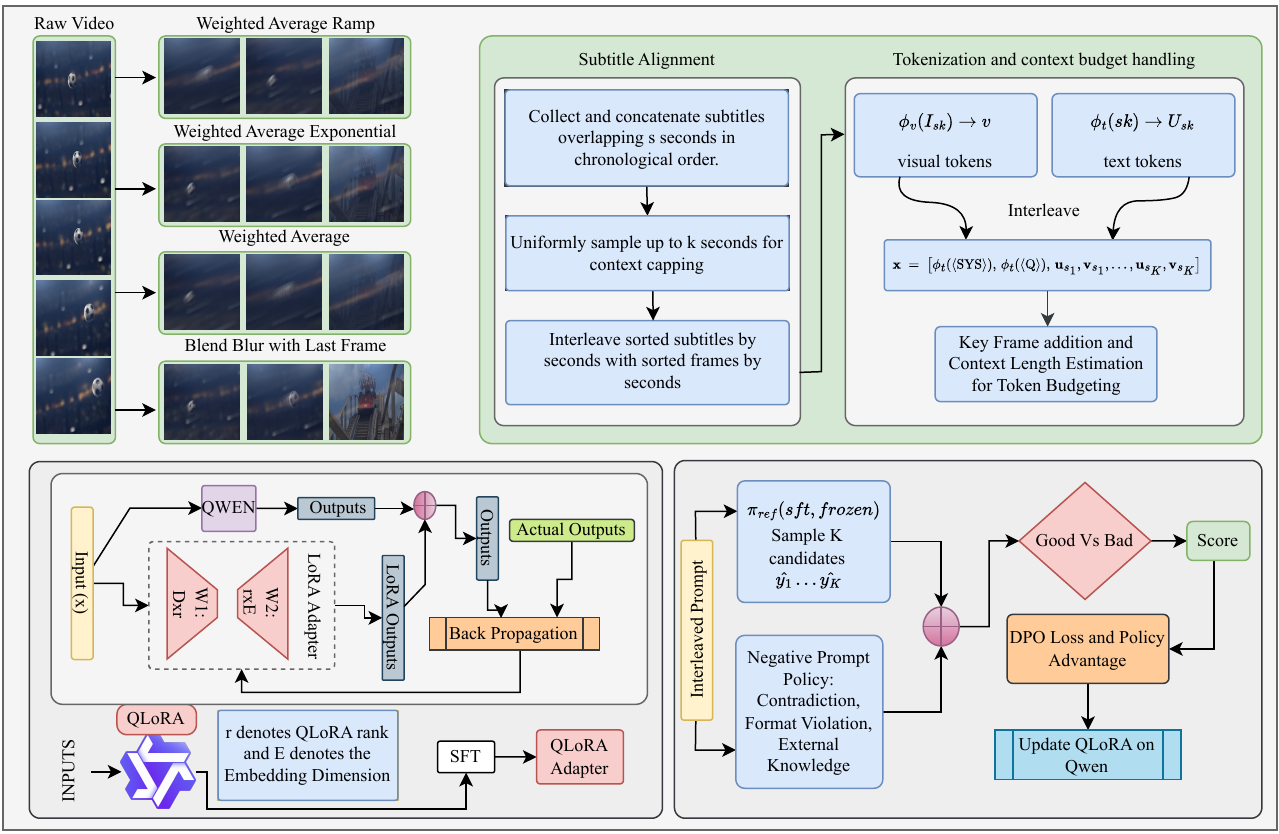}
    \caption{Overview of POVQA. (a) Video is compressed to 1 fps pooled images; (b) SFT trains rationale and final-answer generation; (c) optional DPO aligns preference pairs. The temporal poolers differ in how they weight frames within each one-second window: WA averages frames uniformly within the second, WAE emphasizes more recent frames exponentially, WAR increases recency emphasis with a ramp, and BBLF combines blurred pooled context with the final frame.}
    \label{fig:methodology}
\end{figure*}

\section{Related Work}

\subsection{Long-video understanding and benchmarks}
Video question answering (VQA) benchmarks such as TVQA and MovieQA emphasize multi-step temporal reasoning across long narrative content~\cite{lit1,lei2018tvqa,tapaswi2016movieqa}. Newer datasets (e.g., NExT-QA, STAR, AGQA, AVSD) further stress compositional and temporal understanding~\cite{audio,agqa,star,NExT-QA}. Works such as KnowIT VQA and retrieval-augmented approaches further highlight dialogue and external-knowledge dependencies~\cite{knowit,reveal}. These benchmarks show that long-range evidence integration remains a bottleneck.

\subsection{Video-LVLMs and multimodal instruction tuning}
Recent LVLMs and video-language systems, including Flamingo, PaLI, Qwen-VL variants, Video-LLaVA, TimeChat, and MovieChat, improve general multimodal reasoning~\cite{alayrac2022flamingo,chen2023pali,llava,Maaz2023VideoChatGPT,ren2024timechat,song2024moviechat,yang2025qwen2}. Stronger baselines such as LLaVA-Video and Video-LLaMA families are important comparison points for future large-scale evaluation. In this work we use Qwen2.5-VL-7B because it is open-weight, reproducible, and strong enough to test compression and supervision choices.

\subsection{Temporal modeling, frame sampling, and pooling}
Prior work studies keyframe selection, hierarchical grouping, memory modules, and efficient transformers to reduce video tokens~\cite{vivit,lit4,lit5,action,fastvlm,LongVLM,AdaFrame,lit2,lit3}. Pooling and sampling approaches show that token budget is central to long-video performance, but can lose fine-grained timing signals when compression is too aggressive.

\subsection{Preference optimization and alignment}
Preference-based alignment methods such as DPO have been effective in language and multimodal settings for shaping response behavior~\cite{rafailov2023direct}. Rationale-structured reasoning approaches are also increasingly studied for interpretability~\cite{liu2025commonsense,songmodularized}. In small-data regimes, however, preference supervision can be noisy and sensitive to pair quality, motivating cautious interpretation.

\subsection{Efficiency, test-time constraints, and diagnostics}
Recent work increasingly emphasizes inference-time constraints, token efficiency, and systematic stress-testing of reasoning systems~\cite{lit5,fastvlm}. Our scope is aligned with this direction: we study lightweight evidence compression plus rationale alignment under limited annotation, and evaluate with explicit failure-mode analysis rather than broad benchmark claims.

Unlike large-benchmark works, we target practical long-video VQA under strict context budgets: lightweight compression can cover more seconds, while rationale supervision can improve answer consistency in a controlled pilot setting. We therefore position this as a diagnostics-first study of multimodal algorithmic reasoning under resource constraints. Unlike feature-level temporal modeling methods~\cite{jiang2025storm}, we use a simple image-space pooling step before the LVLM to preserve second-level coverage under a fixed token budget.

\section{POVQA}
POVQA takes a video and subtitle stream, pools raw frames into 1 fps pooled images, interleaves pooled images with subtitle text, and prompts an LVLM to generate a rationale and final answer. Training uses supervised fine-tuning (SFT) on rationale+answer targets, followed by optional Direct Preference Optimization (DPO). Evaluation reports lexical and embedding metrics on free-form outputs.

\subsection{Notation}
Let the video be $V=\{F_{t,k}\}$, where $t$ indexes seconds and $k\in\{1,\ldots,K\}$ indexes frames within second $t$. We denote the pooled image for second $t$ as $I_t$, the pooling weights as $w_k$ with $\sum_{k=1}^{K} w_k = 1$, and the number of pooled visual tokens as $T$ (approximately the video duration in seconds after truncation). We represent subtitle spans as $\mathcal{S}=\{(a_j,b_j,\text{text}_j)\}_{j=1}^{J}$, where $a_j$ and $b_j$ denote the start and end times of subtitle segment $j$, respectively, and $\text{text}_j$ denotes its text content.

For consistency with the equations below, we also use $I_\tau$ for the original frame sequence at $f$ fps, where one-second windows define pooling groups:
\begin{equation}
W_s=\left\{\, I_\tau \mid \tau\in[(s-1)f+1,\; sf] \,\right\}.
\label{eq:window}
\end{equation}
We denote the image encoder by $\phi_v(\cdot)$ (image $\rightarrow$ visual tokens), the text tokenizer/embeddings by $\phi_t(\cdot)$, and the LVLM by $\pi_\theta(\cdot)$.

\subsection{Temporal pooling and pooled frame construction}
For each second $s$, choose nonnegative weights $w_s(\tau)$ on $W_s$ with $\sum_{\tau\in W_s} w_s(\tau)=1$, and form an average
\begin{equation}
\bar{I}_s=\sum_{\tau\in W_s} w_s(\tau)\, I_\tau.
\label{eq:avg}
\end{equation}
Let $G_\sigma(\cdot)$ be a Gaussian blur and $I^{\text{last}}_s = I_{sf}$ the last frame in the second. We instantiate four operators:

\noindent\textbf{Weighted Average (WA)}
\begin{equation}
w_s(\tau)=\frac{1}{|W_s|}, \qquad \tilde{I}_s=\bar{I}_s.
\label{eq:uniform}
\end{equation}

\noindent\textbf{Weighted Average Exponential (WAE): recency bias}
\begin{equation}
w_s(\tau)=\frac{\exp\!\big(\lambda(\tau-sf)\big)}{\sum\limits_{\kappa\in W_s}\exp\!\big(\lambda(\kappa-sf)\big)}, \quad \lambda>0, \qquad \tilde{I}_s=\bar{I}_s.
\label{eq:exp}
\end{equation}

\noindent\textbf{Weighted Average Ramp (WAR): linear recency}
\begin{equation}
w_s(\tau)=\frac{\tau-(s-1)f}{\sum\limits_{\kappa\in W_s}\big(\kappa-(s-1)f\big)}, \qquad \tilde{I}_s=\bar{I}_s.
\label{eq:ramp}
\end{equation}

\noindent\textbf{Blend--Blur with Last Frame (BBLF)}
\begin{equation}
\tilde{I}_s = \alpha\, I^{\text{last}}_s + (1-\alpha)\, G_\sigma\!\big(\bar{I}_s\big), \qquad \alpha\in[0,1],\ \sigma>0.
\label{eq:blendblur}
\end{equation}

Each $\tilde{I}_s$ summarizes motion/appearance from 24--60 raw frames (depending on $f$), compressing intra-second dynamics into one image suitable for tokenization. This reduces visual sequence length from $O(\text{frames})$ to $O(\text{seconds})$.

Pooling keeps dense second-level coverage under a fixed context budget and can preserve broad ``what changed'' signals across scenes. The main failure mode is order-sensitive micro-actions (e.g., fast interactions or tiny objects), where second-level aggregation may blur or reorder crucial evidence.

\subsection{Subtitle alignment and interleaving}
Collect subtitles overlapping second $s$:
\begin{equation}
U_s=\bigoplus_{j:\,[a_j,b_j)\cap[(s-1),s)\neq\emptyset} \text{text}_j,
\label{eq:subs}
\end{equation}
where $\bigoplus$ concatenates spans in chronological order. Cap the context to $S_{\max}$ seconds; if $S>S_{\max}$, uniformly subsample an index set
\begin{equation}
\mathcal{I}\subset\{1,\dots,S\}, \qquad |\mathcal{I}|=K=\min(S,S_{\max}),
\label{eq:subset}
\end{equation}
and sort $\{s_k\}_{k=1}^{K}=\mathrm{sorted}(\mathcal{I})$. Build an interleaved sequence of subtitle spans and pooled images:
\begin{equation}
\mathcal{Z}=\big[\, U_{s_1},\, \tilde{I}_{s_1},\, U_{s_2},\, \tilde{I}_{s_2},\,\dots,\,U_{s_K},\,\tilde{I}_{s_K}\,\big].
\label{eq:interleave}
\end{equation}

\subsection{Tokenization and model input}
Map images and text to tokens:
\begin{equation}
\mathbf{v}_{s_k}=\phi_v(\tilde{I}_{s_k})\in\mathbb{R}^{m\times d}, \qquad \mathbf{u}_{s_k}=\phi_t(U_{s_k}),
\label{eq:tokens}
\end{equation}
and form the full input
\begin{equation}
\mathbf{x}=\big[\phi_t(\langle\mathrm{SYS}\rangle),\ \phi_t(\langle\mathrm{Q}\rangle),\ \mathbf{u}_{s_1},\mathbf{v}_{s_1},\ldots,\mathbf{u}_{s_K},\mathbf{v}_{s_K}\big].
\label{eq:x}
\end{equation}

\subsection{Prompting and rationale format}
We elicit a rationale $\mathbf{y}^{(R)}$ (``Reasoning:'') and a short answer $\mathbf{y}^{(A)}$ (``Final Answer:''). The SYS prompt also includes 1 raw frame of the exact second the user paused the video in our player tool to ask question (if available). We call this frame as the key-frame of question.

We train the model to output rationale and answer in two fields: \texttt{Reasoning: <grounded evidence>} and \texttt{Final Answer: <short answer>}. Rationales are explicit supervised targets during SFT and are part of preference pairs for DPO. Final-answer metrics (F1/BLEU/ROUGE) are computed on the Final Answer field; rationale similarity metrics are reported separately.

\subsection{Supervised fine-tuning (SFT) with QLoRA}
Given supervision $\mathbf{y}=[\mathbf{y}^{(R)},\mathbf{y}^{(A)}]$, the SFT loss is
\begin{equation}
\mathcal{L}_{\mathrm{SFT}}(\theta)=-\mathbb{E}_{(\mathbf{x},\mathbf{y})}\sum_{i=1}^{|\mathbf{y}|}\log \pi_\theta\!\big(y_i \mid \mathbf{x}, \mathbf{y}_{<i}\big).
\label{eq:sft}
\end{equation}

We implement SFT with QLoRA adapters for parameter-efficient tuning under limited hardware~\cite{dettmers2023qlora}. For a pretrained matrix $W\in\mathbb{R}^{d_{\mathrm{out}}\times d_{\mathrm{in}}}$, QLoRA keeps a quantized frozen copy $q(W)$ and learns a low-rank update:
\begin{equation}
\begin{aligned}
\Delta W &= \frac{\alpha}{r}BA,\\
A &\in \mathbb{R}^{r\times d_{\mathrm{in}}},\\
B &\in \mathbb{R}^{d_{\mathrm{out}}\times r},\\
r &\ll \min(d_{\mathrm{in}},d_{\mathrm{out}}).
\end{aligned}
\label{eq:lora2}
\end{equation}
Only $A,B$ (and selected norms/biases) are trained under Eq.~\eqref{eq:sft}.

\subsection{Direct Preference Optimization (DPO)}
Let $\mathcal{D}_{\mathrm{pref}}=\{(\mathbf{x},\mathbf{y}^{+},\mathbf{y}^{-})\}$ be preference triples (preferred vs.\ dispreferred rationale+answer). With a frozen reference policy $\pi_{\mathrm{ref}}$ (the SFT model) and current policy $\pi_\theta$, DPO minimizes
\begin{equation}
\begin{aligned}
\Delta(\mathbf{x},\mathbf{y}^{+},\mathbf{y}^{-})=\;
&\big[\log\pi_\theta(\mathbf{y}^{+}\mid\mathbf{x})-\log\pi_\theta(\mathbf{y}^{-}\mid\mathbf{x})\big]\\
&-\big[\log\pi_{\mathrm{ref}}(\mathbf{y}^{+}\mid\mathbf{x})-\log\pi_{\mathrm{ref}}(\mathbf{y}^{-}\mid\mathbf{x})\big].
\end{aligned}
\label{eq:dpo}
\end{equation}
where $\beta>0$ and $\sigma(\cdot)$ is the logistic sigmoid. Sequence log-likelihoods expand tokenwise:
\begin{equation}
\log\pi_\theta(\mathbf{y}\mid\mathbf{x})=\sum_{i=1}^{|\mathbf{y}|}\log \pi_\theta\!\big(y_i \mid \mathbf{x}, \mathbf{y}_{<i}\big).
\label{eq:seqll}
\end{equation}
Given Eq.~\eqref{eq:dpo}, DPO minimizes the logistic loss
\begin{equation}
\mathcal{L}_{\mathrm{DPO}}(\theta)=-\,\mathbb{E}_{(\mathbf{x},\mathbf{y}^{+},\mathbf{y}^{-})}\big[\log \sigma\!\big(\beta\,\Delta(\mathbf{x},\mathbf{y}^{+},\mathbf{y}^{-})\big)\big]
\label{eq:dpo_loss}
\end{equation}
where $\sigma(\cdot)$ is the logistic sigmoid and $\beta>0$. In our setting, DPO is optional and analyzed cautiously: it can align responses toward preferred outputs, but small preference sets can introduce instability or metric tradeoffs.

\subsection{Token budget and attention shift}
Each pooled image yields $m$ visual tokens and subtitles add $|\mathbf{u}_{s_k}|$ text tokens. For $K\leq S_{\max}$ seconds the context is
\begin{equation}
N_{\mathrm{ctx}} \approx N_{\mathrm{sys+Q}}+\sum_{k=1}^{K}(|\mathbf{u}_{s_k}|+m),
\label{eq:budget}
\end{equation}
independent of raw fps $f$ since pooling maps $\Theta(f)$ frames to $m$ tokens/second. Thus motion is absorbed in $\tilde{I}_s$ by $\phi_v$, while temporal attention operates over adjacent pooled seconds. This design is related to prior work on segment-level temporal aggregation and video token efficiency~\cite{attention,tong2022videomae,tsn}.

For a 5-minute clip at 24 fps with $S_{\max}=60$, $m=256$, average subtitle length 10 tokens/second, and $N_{\mathrm{sys+Q}}=128$, pooled input is about 16{,}088 tokens versus about 369{,}368 tokens without pooling over the same 60-second budget. This is about a $23\times$ context reduction while preserving second-level temporal coverage.

\subsection{Decoding}
At inference we reuse Eq.~\eqref{eq:interleave}--Eq.~\eqref{eq:x} and decode sequentially with $\mathcal{C}$ as the answer span regulator:
\begin{align}
\hat{\mathbf{y}}^{(R)} &= \arg\max_{\mathbf{y}} \pi_\theta(\mathbf{y}\mid \mathbf{x}, \text{``Reasoning:''}), \\
\hat{\mathbf{y}}^{(A)} &= \arg\max_{\mathbf{y}\in\mathcal{C}} \pi_\theta(\mathbf{y}\mid \mathbf{x}, \hat{\mathbf{y}}^{(R)}, \text{``Final Answer:''}).
\end{align}

\section{Experiments}
\begin{table}[!t]
\centering
\small
\setlength{\tabcolsep}{3pt}
\renewcommand{\arraystretch}{1.05}
\caption{Transposed base results on ReasonVQA eval. B1/B4: BLEU-1/4, RL: ROUGE-L, EC: Embed Cos, RLR: ROUGE-L-R, ECR: Embed Cos-R. Highlights: \colorbox{high1}{highest}, \colorbox{high2}{second-highest}.}
\label{tab:base_results}
\begin{tabular}{lccccc}
\toprule
Metric & BBLF & WA & WAE & WAR & KF \\
\midrule
F1  & \cellcolor{high1}0.212 & \cellcolor{high2}0.204 & \cellcolor{high2}0.204 & 0.184 & 0.070 \\
B1  & \cellcolor{high1}0.453 & \cellcolor{high2}0.441 & 0.405 & 0.380 & 0.159 \\
B4  & 0.021 & \cellcolor{high1}0.031 & \cellcolor{high2}0.023 & 0.021 & 0.000 \\
RL  & \cellcolor{high1}0.196 & \cellcolor{high2}0.187 & 0.183 & 0.168 & 0.069 \\
EC  & \cellcolor{high1}0.383 & \cellcolor{high2}0.365 & 0.363 & 0.361 & 0.197 \\
RLR & 0.172 & 0.167 & \cellcolor{high1}0.177 & \cellcolor{high2}0.173 & 0.165 \\
ECR & 0.533 & 0.532 & \cellcolor{high2}0.548 & \cellcolor{high1}0.551 & 0.533 \\
\bottomrule
\end{tabular}
\end{table}
\subsection{Experimental Protocol}

\subsubsection{Dataset and splits}
We use ReasonVQA, a pilot diagnostic dataset with 239 question--answer--rationale triplets from 12 movies (romance, historical, biography, western, fantasy, action, mystery, thriller, animation, drama, sci-fi, documentary). The evaluation split contains held-out sci-fi and western titles. Across all questions, the source context spans over 1M raw frames. This scale is intentionally small and is used for controlled ablation, not broad benchmark claims.

\subsubsection{Input budget and decoding}
Unless otherwise stated, we pool video to 1 fps pooled images, cap context to 60 pooled seconds, and interleave pooled images with subtitle text. For generation we decode rationale plus final answer with the same prompt format used in training.

\subsubsection{Model and training}
We adopt Qwen2.5-VL-7B for open weights and stable tooling. Our main hyperparameters are QLoRA rank/$\alpha$=32, dropout=0.05, seed=42, grad-acc=8, learning rate 5e-5 (SFT) and 5e-6 (DPO). We append a key-frame instruction in the system prompt and uniformly sample 16 pooled frames per training step. Training used one NVIDIA A40-48Q (48 GB).

\subsubsection{Baselines and fairness}
We report three main settings: pooled-only (no fine-tuning), pooled+SFT, and pooled+SFT+DPO. We also include a keyframe-only ablation. For TVQA, our setting is zero-shot transfer from ReasonVQA; this is not directly comparable to fully supervised TVQA systems trained on TVQA labels.

\subsubsection{Metrics and their limitations}
We report F1, BLEU, and ROUGE-L for final answers, plus embedding-similarity and overlap metrics for rationales. These metrics are useful for large ablations but do not perfectly track human judgment. We therefore include qualitative examples and a failure taxonomy. Unless noted, results are from single runs and variance is not reported. This evaluation protocol is intentionally diagnostic: we focus on systematic strengths and failure modes in multimodal reasoning under fixed test-time budgets.

\begin{table}[!t]
\centering
\small
\setlength{\tabcolsep}{4pt}
\renewcommand{\arraystretch}{1}
\caption{Cross-evaluation of fine-tuned models after SFT in ReasonVQA eval set. Highlights: \colorbox{high1}{highest}, \colorbox{high2}{second-highest}, \colorbox{gray!15}{same training and evaluation method}. Method abbreviations: BBLF (Blend Blur Last Frame), WA (Weighted Avg), WAE (Weighted Avg Exp), WAR (Weighted Avg Ramp), Metric-R (Metric- Reasoning)}
\label{tab:sft_cross_eval_fancy}
\begin{tabular}{lcccc}
\toprule
& \multicolumn{4}{c}{\textbf{Model Trained On}} \\
\cmidrule(l){2-5}
Metric & \makecell[ct]{BBLF} & \makecell[ct]{WA} & \makecell[ct]{WAE} & \makecell[ct]{WAR} \\
\midrule
\multicolumn{5}{l}{\textit{Evaluated on: Blend Blur With Last Frame}} \\
\midrule
F1& \cellcolor{gray!15} 0.521 & 0.468 & \cellcolor{high1}0.543 & \cellcolor{high2}0.525 \\
BLEU-1& \cellcolor{gray!15} 0.574 & 0.520 & \cellcolor{high1}0.603 & \cellcolor{high2}0.581 \\
BLEU-4 (BP)& \cellcolor{gray!15} 0.245 & 0.209 & \cellcolor{high1}0.265 & \cellcolor{high2}0.257 \\
ROUGE-L& \cellcolor{gray!15} 0.499 & 0.445 & \cellcolor{high1}0.520 & \cellcolor{high2}0.504 \\
Embed Cosine& \cellcolor{gray!15} \cellcolor{high2}0.620 & 0.573 & \cellcolor{high1}0.632 & 0.617 \\
ROUGE-L-R& \cellcolor{gray!15} \cellcolor{high2}0.227 & 0.206 & \cellcolor{high1}0.227 & 0.225 \\
Embed Cosine-R& \cellcolor{gray!15} 0.586 & 0.583 & \cellcolor{high1}0.590 & \cellcolor{high2}0.587 \\
\midrule
\multicolumn{5}{l}{\textit{Evaluated on: Weighted Average}} \\
\midrule
F1& 0.521 & \cellcolor{gray!15} 0.523 & \cellcolor{high1}0.550 & \cellcolor{high2}0.535 \\
BLEU-1& 0.580 & \cellcolor{gray!15} 0.580 & \cellcolor{high1}0.604 & \cellcolor{high2}0.587 \\
BLEU-4 (BP)& 0.237 & \cellcolor{gray!15} 0.226 & \cellcolor{high1}0.278 & \cellcolor{high2}0.256 \\
ROUGE-L& 0.495 & \cellcolor{gray!15} 0.492 & \cellcolor{high1}0.520 & \cellcolor{high2}0.503 \\
Embed Cosine& 0.620 & \cellcolor{gray!15} 0.622 & \cellcolor{high1}0.627 & \cellcolor{high2}0.627 \\
ROUGE-L-R& 0.241 & \cellcolor{gray!15} \cellcolor{high2}0.244 & 0.238 & \cellcolor{high1}0.245 \\
Embed Cosine-R& \cellcolor{high1}0.592 & \cellcolor{gray!15} 0.588 & 0.589 & \cellcolor{high2}0.590 \\
\midrule
\multicolumn{5}{l}{\textit{Evaluated on: Weighted Average (Exp)}} \\
\midrule
F1& \cellcolor{high1}0.533 & 0.520 & \cellcolor{gray!15} 0.506 & \cellcolor{high2}0.520 \\
BLEU-1& \cellcolor{high1}0.583 & \cellcolor{high2}0.577 & \cellcolor{gray!15} 0.553 & 0.574 \\
BLEU-4 (BP)& \cellcolor{high1}0.248 & \cellcolor{high2}0.230 & \cellcolor{gray!15} 0.217 & 0.214 \\
ROUGE-L& \cellcolor{high1}0.504 & \cellcolor{high2}0.496 & \cellcolor{gray!15} 0.477 & 0.491 \\
Embed Cosine& 0.600 & \cellcolor{high2}0.609 & \cellcolor{gray!15} 0.588 & \cellcolor{high1}0.614 \\
ROUGE-L-R& 0.230 & 0.224 & \cellcolor{gray!15} \cellcolor{high2}0.230 & \cellcolor{high1}0.231 \\
Embed Cosine-R& 0.574 & \cellcolor{high2}0.581 & \cellcolor{gray!15} 0.572 & \cellcolor{high1}0.587 \\
\midrule
\multicolumn{5}{l}{\textit{Evaluated on: Weighted Average (Ramp)}} \\
\midrule
F1& 0.524 & \cellcolor{high1}0.545 & 0.519 & \cellcolor{gray!15} \cellcolor{high2}0.526 \\
BLEU-1& 0.575 & \cellcolor{high1}0.603 & 0.572 & \cellcolor{gray!15} \cellcolor{high2}0.586 \\
BLEU-4 (BP)& 0.224 & \cellcolor{high1}0.247 & 0.216 & \cellcolor{gray!15} \cellcolor{high2}0.228 \\
ROUGE-L& 0.490 & \cellcolor{high1}0.512 & 0.488 & \cellcolor{gray!15} \cellcolor{high2}0.495 \\
Embed Cosine& \cellcolor{high2}0.622 & \cellcolor{high1}0.630 & 0.605 & \cellcolor{gray!15} 0.620 \\
ROUGE-L-R& 0.236 & \cellcolor{high1}0.243 & \cellcolor{high2}0.242 & \cellcolor{gray!15} 0.241 \\
Embed Cosine-R& \cellcolor{high1}0.597 & \cellcolor{high2}0.596 & 0.596 & \cellcolor{gray!15} 0.592 \\
\bottomrule
\end{tabular}
\end{table}

\begin{table}[!t]
\centering
\small
\setlength{\tabcolsep}{4pt}
\renewcommand{\arraystretch}{1}
\caption{Cross-evaluation of fine-tuned models after SFT + DPO in ReasonVQA eval set. Highlights: \colorbox{high1}{highest}, \colorbox{high2}{second-highest}, \colorbox{gray!15}{same training and evaluation method}. Method abbreviations: BBLF (Blend Blur Last Frame), WA (Weighted Avg), WAE (Weighted Avg Exp), WAR (Weighted Avg Ramp), Metric-R (Metric- Reasoning)}
\label{tab:dpo_cross_eval_fancy}
\begin{tabular}{lcccc}
\toprule
& \multicolumn{4}{c}{\textbf{Model Trained On}} \\
\cmidrule(l){2-5}
Metric & \makecell[ct]{BBLF} & \makecell[ct]{WA} & \makecell[ct]{WAE} & \makecell[ct]{WAR} \\
\midrule
\multicolumn{5}{l}{\textit{Evaluated on: Blend Blur With Last Frame}} \\
\midrule
F1& \cellcolor{gray!15} 0.505 & \cellcolor{high2}0.506 & \cellcolor{high1}0.541 & 0.495 \\
BLEU-1& \cellcolor{gray!15} 0.553 & \cellcolor{high2}0.558 & \cellcolor{high1}0.583 & 0.543 \\
BLEU-4 (BP)& \cellcolor{gray!15} 0.231 & 0.218 & \cellcolor{high1}0.267 & \cellcolor{high2}0.233 \\
ROUGE-L& \cellcolor{gray!15} \cellcolor{high2}0.484 & 0.482 & \cellcolor{high1}0.513 & 0.474 \\
Embed Cosine& \cellcolor{gray!15} 0.614 & \cellcolor{high2}0.614 & \cellcolor{high1}0.631 & 0.592 \\
ROUGE-L-R& \cellcolor{gray!15} \cellcolor{high2}0.230 & 0.224 & \cellcolor{high1}0.233 & 0.224 \\
Embed Cosine-R& \cellcolor{gray!15} \cellcolor{high2}0.588 & 0.588 & \cellcolor{high1}0.597 & 0.587 \\
\midrule
\multicolumn{5}{l}{\textit{Evaluated on: Weighted Average}} \\
\midrule
F1& 0.518 & \cellcolor{gray!15} \cellcolor{high2}0.527 & \cellcolor{high1}0.541 & 0.483 \\
BLEU-1& 0.573 & \cellcolor{gray!15} \cellcolor{high2}0.580 & \cellcolor{high1}0.594 & 0.531 \\
BLEU-4 (BP)& \cellcolor{high2}0.254 & \cellcolor{gray!15} 0.246 & \cellcolor{high1}0.272 & 0.217 \\
ROUGE-L& 0.496 & \cellcolor{gray!15} \cellcolor{high2}0.501 & \cellcolor{high1}0.518 & 0.459 \\
Embed Cosine& 0.605 & \cellcolor{gray!15} \cellcolor{high1}0.619 & \cellcolor{high2}0.610 & 0.575 \\
ROUGE-L-R& 0.228 & \cellcolor{gray!15} \cellcolor{high1}0.236 & \cellcolor{high2}0.229 & 0.227 \\
Embed Cosine-R& 0.585 & \cellcolor{gray!15} 0.586 & \cellcolor{high2}0.586 & \cellcolor{high1}0.587 \\
\midrule
\multicolumn{5}{l}{\textit{Evaluated on: Weighted Average (Exp)}} \\
\midrule
F1& \cellcolor{high1}0.526 & \cellcolor{high2}0.521 & \cellcolor{gray!15} 0.505 & 0.514 \\
BLEU-1& \cellcolor{high2}0.568 & \cellcolor{high1}0.575 & \cellcolor{gray!15} 0.560 & 0.562 \\
BLEU-4 (BP)& \cellcolor{high1}0.240 & \cellcolor{high2}0.226 & \cellcolor{gray!15} 0.212 & 0.210 \\
ROUGE-L& \cellcolor{high1}0.498 & \cellcolor{high2}0.498 & \cellcolor{gray!15} 0.479 & 0.489 \\
Embed Cosine& \cellcolor{high2}0.602 & \cellcolor{high1}0.608 & \cellcolor{gray!15} 0.577 & 0.599 \\
ROUGE-L-R& \cellcolor{high1}0.234 & \cellcolor{high2}0.231 & \cellcolor{gray!15} 0.226 & 0.229 \\
Embed Cosine-R& \cellcolor{high2}0.587 & \cellcolor{high1}0.592 & \cellcolor{gray!15} 0.574 & 0.586 \\
\midrule
\multicolumn{5}{l}{\textit{Evaluated on: Weighted Average (Ramp)}} \\
\midrule
F1& \cellcolor{high1}0.543 & 0.535 & 0.512 & \cellcolor{gray!15} \cellcolor{high2}0.541 \\
BLEU-1& \cellcolor{high1}0.602 & 0.597 & 0.571 & \cellcolor{gray!15} \cellcolor{high2}0.600 \\
BLEU-4 (BP)& \cellcolor{high1}0.273 & 0.230 & 0.233 & \cellcolor{gray!15} \cellcolor{high2}0.252 \\
ROUGE-L& \cellcolor{high1}0.516 & 0.502 & 0.483 & \cellcolor{gray!15} \cellcolor{high2}0.512 \\
Embed Cosine& \cellcolor{high2}0.624 & 0.621 & 0.585 & \cellcolor{gray!15} \cellcolor{high1}0.629 \\
ROUGE-L-R& \cellcolor{high2}0.243 & 0.242 & 0.232 & \cellcolor{gray!15} \cellcolor{high1}0.246 \\
Embed Cosine-R& \cellcolor{high2}0.593 & \cellcolor{high1}0.594 & 0.591 & \cellcolor{gray!15} 0.589 \\
\bottomrule
\end{tabular}
\end{table}

\subsection{Results}

\subsubsection{ReasonVQA}
Table~\ref{tab:base_results} reports pooled-only baselines before adaptation. Baseline answer quality is low (best F1 0.212). After two epochs of SFT, cross-evaluation results in Table~\ref{tab:sft_cross_eval_fancy} reach up to 0.550 F1 depending on train/eval pooling pairing; with SFT+DPO (Table~\ref{tab:dpo_cross_eval_fancy}), the best F1 is 0.543. Relative to pooled-only, SFT provides the main lift in this pilot setting. Rationale similarity also increases, e.g., Embed Cosine-R from 0.533 (base) to 0.597 (best DPO setting).

Table~\ref{tab:dpo_delta_by_eval} shows that DPO effects are mixed: some rationale-oriented metrics improve, while answer-overlap metrics can be flat or slightly lower depending on the pooler. This is consistent with small-scale preference optimization, where pair noise and style alignment can trade off against lexical overlap.

This pattern matches DPO's objective, which optimizes preference consistency rather than directly maximizing BLEU/ROUGE/F1. Overall, SFT drives most gains, while DPO is best interpreted as an optional alignment stage.

We can also observe from Tab.~\ref{tab:dpo_cross_eval_fancy} and Tab.~\ref{tab:sft_cross_eval_fancy} that the diagonal is not always optimal. Under WAE evaluation, the best SFT model is cross-trained on BBLF (F1 = 0.533) rather than the diagonal (0.506). Under WAR evaluation, SFT's best is the WA-trained model (F1 = 0.545) rather than the diagonal (0.526), whereas after DPO BBLF and WAE perform most consistently. This indicates that training on appearance-preserving pooling (BBLF) transfers well to motion-sensitive evaluation. This suggests that cross-pooler transfer may reflect learning temporally robust evidence patterns rather than overfitting to a single pooling operator. In this sense, the gains appear to come from transferable second-level cues, not only exact train--test pooler matching.

\subsubsection{Zero-shot transfer to TVQA}
We evaluate POVQA on the TVQA eval set in a strict zero-shot setting after fine tuning on ReasonVQA.

The only change to the pipeline in Fig.~\ref{fig:methodology} is the system prompt to expose the answer options. We sample a random 5k subset for the evaluation split (scripts provided) for computational reason with $p \equiv 0.64$ and $n = 5000$ the normal approximated 95\% CI is about $\pm 1.4$ points.

POVQA (SFT+DPO) attains 64.7\% zero-shot accuracy on TVQA~\cite{lei2018tvqa}, while pooling-only attains 69.7\% (Tabs.~\ref{tab:tvqa_all} and~\ref{tab:tvqa_base_halfcol}). Because many prior systems in Table~\ref{tab:tvqa_all} are trained under different supervision or modality settings, we treat this comparison as contextual rather than a like-for-like leaderboard claim.

Table~\ref{tab:sft_cross_eval_fancy} indicates that SFT consistently improves over pooled-only baselines and that cross-pooler transfer is often stronger than diagonal-only training.

Table~\ref{tab:dpo_cross_eval_fancy} shows that DPO is competitive but not uniformly better than SFT-only on answer overlap, reinforcing cautious interpretation in small-data settings.

\subsubsection{DPO analysis}
From Table~\ref{tab:dpo_delta_by_eval}, DPO tends to help more on rationale-oriented metrics than on strict lexical answer metrics. A plausible explanation is preference-pair noise and overfitting in a limited corpus: for ambiguous questions, DPO can improve concise preference alignment, while for precise temporal ordering it can slightly reduce lexical match.

\begin{table}[!t]
\centering
\footnotesize
\setlength{\tabcolsep}{3pt}
\renewcommand{\arraystretch}{1}
\caption{DPO vs.\ SFT deltas by evaluation pooler (best over training poolers) in ReasonVQA eval set. $\Delta$ = DPO $-$ SFT; positive means DPO helps. Highlights: \colorbox{high1}{highest}, \colorbox{high2}{second-highest}.}
\label{tab:dpo_delta_by_eval}
\begin{tabular}{lrrrr}
\toprule
& \multicolumn{4}{c}{\textbf{Model Trained On}} \\
\cmidrule(l){2-5}
Metric & \makecell[ct]{BBLF} & \makecell[ct]{WA} & \makecell[ct]{WAE} & \makecell[ct]{WAR} \\
\midrule
F1 & \cellcolor{high1}-0.002 & -0.009 & -0.007 & \cellcolor{high2}-0.002 \\
BLEU-1 & -0.020 & -0.010 & \cellcolor{high2}-0.008 & \cellcolor{high1}-0.001 \\
BLEU-4 (BP) & \cellcolor{high2}+0.002 & -0.006 & -0.008 & \cellcolor{high1}+0.026 \\
ROUGE-L & -0.007 & \cellcolor{high2}-0.002 & -0.006 & \cellcolor{high1}+0.004 \\
Embed Cosine & \cellcolor{high1}-0.001 & -0.008 & -0.006 & \cellcolor{high2}-0.001 \\
ROUGE-L-R & \cellcolor{high1}+0.006 & -0.009 & \cellcolor{high2}+0.003 & +0.003 \\
Embed Cosine-R & \cellcolor{high1}+0.007 & -0.005 & \cellcolor{high2}+0.005 & -0.003 \\
\bottomrule
\end{tabular}
\end{table}

\begin{figure*}[!t]
    \centering
    \begin{subfigure}[t]{0.49\linewidth}
        \centering
        \includegraphics[width=\linewidth]{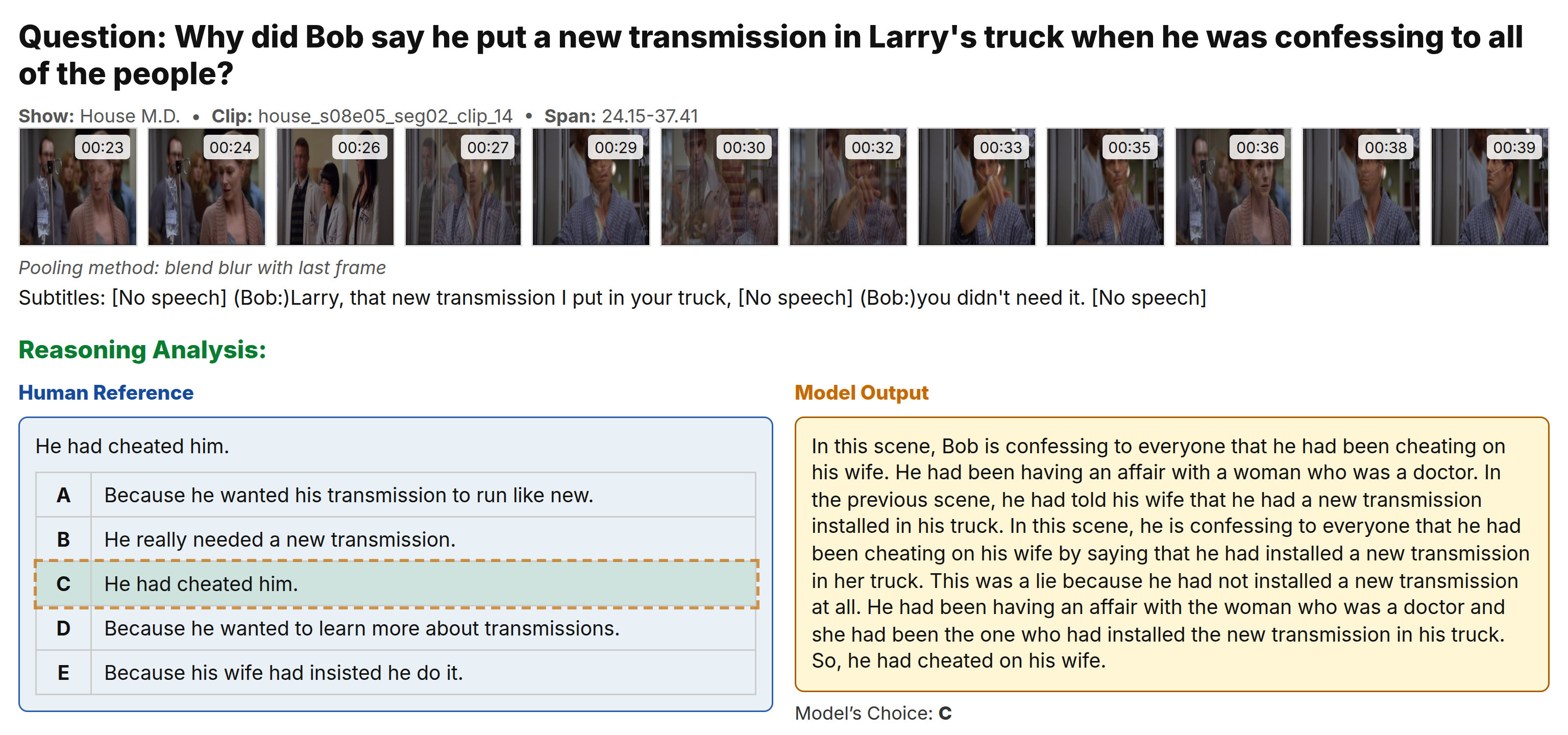}
        \caption{TVQA qualitative samples.}
    \end{subfigure}\hfill
    \begin{subfigure}[t]{0.49\linewidth}
        \centering
        \includegraphics[width=\linewidth]{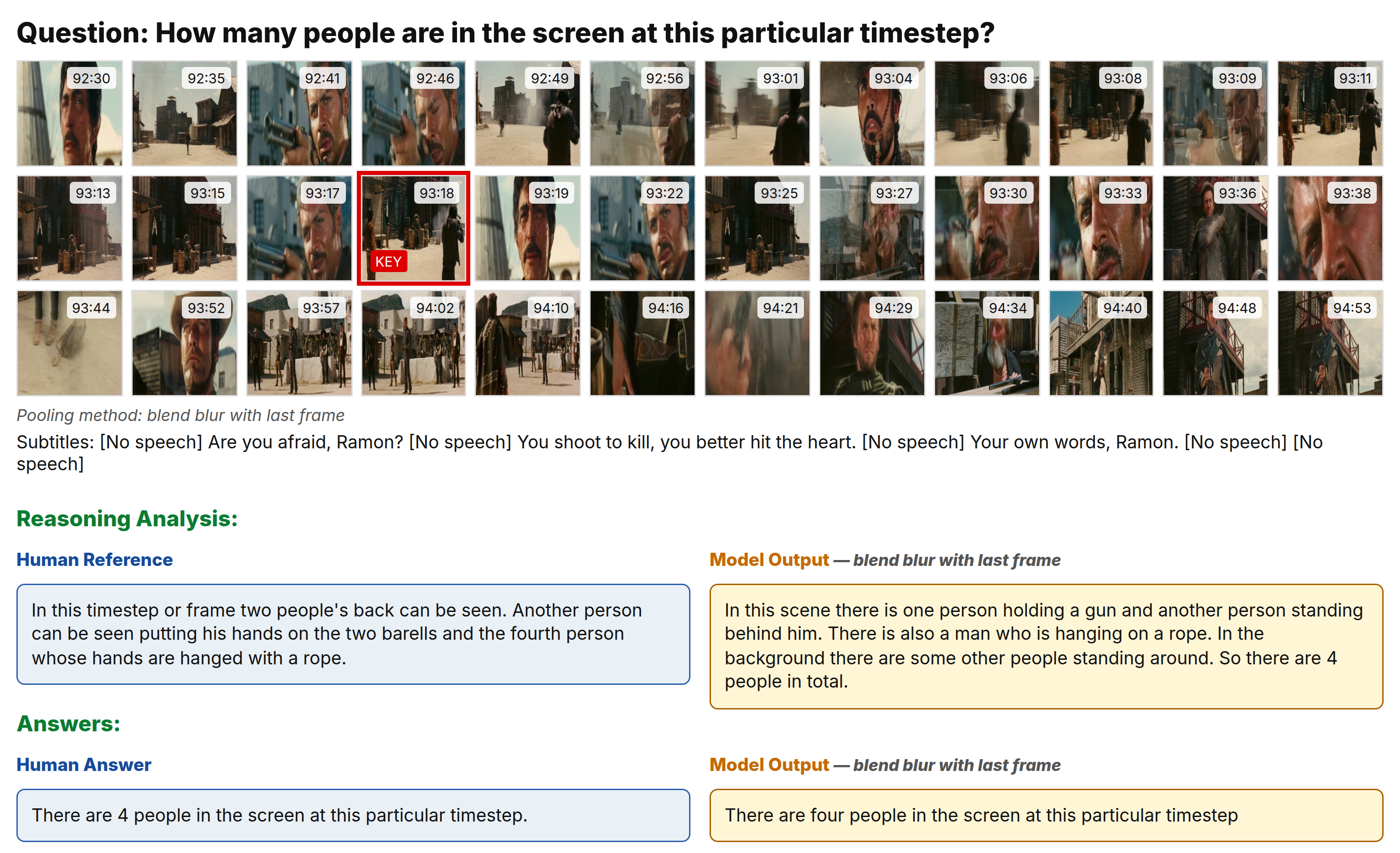}
        \caption{ReasonVQA qualitative samples.}
    \end{subfigure}
    \caption{Qualitative analysis on 2 randomly selected examples. Frames are sub-sampled for readability. Additional qualitative samples, including both successful and failure cases, are provided on the project page: \href{https://povqa.github.io}{https://povqa.github.io}}
    \label{fig:qualitative_analysis_combined}
\end{figure*}
\begin{table}[!t]
\centering
\footnotesize
\setlength{\tabcolsep}{4pt}
\renewcommand{\arraystretch}{1.01}
\caption{TVQA accuracy (\%) for contextual reference only, not as a like-for-like leaderboard comparison. Zero-shot means no TVQA training. ``w/ speech'' denotes the use of ASR/subtitles. Our row reports the best result across evaluation poolers (BBLF/WA/WAE/WAR) with training on BBLF only. Highlights: \colorbox{high1}{our result}, \colorbox{high2}{selected reference results}.}
\label{tab:tvqa_all}
\begin{tabular}{>{\raggedright\arraybackslash}p{0.3\columnwidth} c c c}
\toprule
\textbf{Model} & \textbf{Zero-shot} & \textbf{Venue (Year)} & \textbf{Acc.} \\
\midrule
FrozenBiLM \cite{yang2022zero} & \xmark & NeurIPS (2022) & 82.0 \\
VINDLU \cite{cheng2023vindlu} & \xmark & CVPR (2023) & 79.0 \\
HERO \cite{li-etal-2020-hero} & \xmark & EMNLP (2020) & 74.24 \\
SeViLA \cite{yu2023self} & \xmark & NeurIPS (2023) & \cellcolor{high2}61.6 \\
ViLA \cite{wang2024vila} & \xmark & ECCV (2024) & \cellcolor{high2}63.4 \\
BLIP-2 \cite{li2023blip} & \xmark & ICML (2023) & \cellcolor{high2}54.5 \\
InternVideo \cite{wang2022internvideo} & \xmark & arXiv (2022) & \cellcolor{high2}57.2 \\
\midrule
FrozenBiLM (w/ speech) \cite{yang2022zero} & \cmark & NeurIPS (2022) & \cellcolor{high2}59.7 \\
FrozenBiLM (vision-only) \cite{yang2022zero} & \cmark & NeurIPS (2022) & 29.7 \\
IG-VLM (LLaVA-1.6 34B) \cite{kim2024image} & \cmark & IEEE Access (2024) & 51.1 \\
GPT-4V (via IG-VLM) \cite{kim2024image,achiam2023gpt} & \cmark & arXiv (2024) & 57.8 \\
Goldfish--7B (vision+subs) \cite{ataallah2024goldfish} & \cmark & ECCV (2024) & 46.94 \\
Goldfish--7B (vision only) \cite{ataallah2024goldfish} & \cmark & ECCV (2024) & 36.45 \\
Q-ViD \cite{mogrovejo2024question} & \cmark & Findings ACL (2024) & 41.0 \\
VideoChat2 (reported by Q-ViD) \cite{li2024mvbench} & \cmark & CVPR (2024) & 40.6 \\
SeViLA (reported by Q-ViD) \cite{yu2023self} & \cmark & NeurIPS (2023) & 38.2 \\
InternVideo (reported by Q-ViD) \cite{wang2022internvideo} & \cmark & arXiv (2022) & 35.9 \\
\midrule
\textbf{POVQA (ours)} & \cmark & \textbf{This work (2026)} & \cellcolor{high1}64.7 \\
\bottomrule
\end{tabular}
\end{table}

We highlight only the gap between pooling vs.\ keyframe under identical zero-shot protocol; other rows are for context.

\subsection{Ablation}
Most of our experiments were ablation-driven to isolate the effect of (i) frame-pooling strategy, (ii) supervision objective (SFT vs.\ DPO), and (iii) frame selection (retaining motion-blurred frames). Tab.~\ref{tab:base_results} establishes the lower bound by removing motion-blurred frames and yields among the lowest scores, indicating that even ``imperfect'' frames contribute temporal evidence. Building on that, Tabs.~\ref{tab:base_results} to~\ref{tab:dpo_cross_eval_fancy} report the full cross-evaluation sweep across four pooling strategies: Blend BBLF, WA, WAE, and WAR under both SFT and DPO (37 configurations). Across these multi-frame sweeps, BBLF emerges as the most consistently strong choice after fine-tuning. To probe the strength of single-frame shortcuts against our 60-frame regime, we additionally ran a KeyFrame-only ablation (+8 configurations) and TVQA zero-shot runs bringing the total to 50 experimental sweeps.

\begin{table}[!t]
\centering
\footnotesize
\setlength{\tabcolsep}{3pt}
\renewcommand{\arraystretch}{1}
\caption{KeyFrame-only ablation (max over SFT/DPO per method). \colorbox{high1}{highest}, \colorbox{high2}{second-highest} per row. $\Delta$ = best DPO (over methods) $-$ best KeyFrame ablation (over methods).}
\label{tab:keyframe_ablation_transposed_delta}
\begin{tabular}{lccccr}
\toprule
Metric & \makecell[ct]{BBLF} & \makecell[ct]{WA} & \makecell[ct]{WAE} & \makecell[ct]{WAR} & $\Delta$ \\
\midrule
F1 & \cellcolor{high2} 0.558 & 0.553 & 0.555 & \cellcolor{high1} 0.563 & -0.005 \\
BLEU-1 & 0.618 & 0.616 & \cellcolor{high1} 0.621 & \cellcolor{high2} 0.620 & -0.003 \\
BLEU-4 (BP) & 0.289 & 0.287 & \cellcolor{high1} 0.291 & \cellcolor{high2} 0.291 & +0.000 \\
ROUGE-L & 0.523 & 0.518 & \cellcolor{high2} 0.524 & \cellcolor{high1} 0.528 & -0.007 \\
Embed Cosine & \cellcolor{high1} 0.619 & 0.616 & 0.602 & \cellcolor{high2} 0.617 & +0.013 \\
ROUGE-L-R & 0.240 & \cellcolor{high1} 0.246 & \cellcolor{high2} 0.243 & 0.240 & +0.000 \\
Embed Cos-R & 0.562 & \cellcolor{high1} 0.564 & \cellcolor{high2} 0.562 & 0.561 & +0.033 \\
\bottomrule
\end{tabular}
\end{table}

Tab.~\ref{tab:keyframe_ablation_transposed_delta} shows that KeyFrame-only token metrics cluster tightly (F1 0.553--0.563, ROUGE-L 0.518--0.528, BLEU-4 0.287--0.291), i.e., they are largely insensitive to temporal evidence. The only clearly positive deltas appear on embedding-based metrics: Embed Cosine $\Delta = +0.013$ and Embed Cos-R $\Delta = +0.033$, which capture semantic fidelity that a single frame cannot supply. In contrast, lexical deltas are near-zero or slightly negative (F1 $-0.005$, BLEU-1 $-0.003$, ROUGE-L $-0.007$). Thus, even when token scores look close under KeyFrame-only evaluation, using all 60 frames primarily buys semantic grounding and reasoning consistency, which is exactly what VQA is supposed to test.

\subsection{Qualitative Analysis and Failure Cases}
Figs.~\ref{fig:qualitative_analysis_combined}a and~\ref{fig:qualitative_analysis_combined}b show both successful and failure examples. Success cases typically involve broad scene context or character-state changes that are preserved by second-level pooling. Failure cases mainly follow four patterns: temporal-order errors in fast actions, small object/action misses, dialogue-dependency errors, and occlusion or hard scene cuts. These failure modes match the known limitations of temporal pooling under fixed context budgets.

On TVQA, the zero-shot base model with interleaved pooling surpasses our SFT/DPO adapters, whereas the KeyFrame-only variant lags (Tab.~\ref{tab:tvqa_base_halfcol}). This indicates that temporal pooling is a primary driver of transfer in our setup; preference tuning may require larger, domain-matched preference pairs to improve accuracy consistently.

\begin{table}[!t]
\centering
\footnotesize
\setlength{\tabcolsep}{6pt}
\renewcommand{\arraystretch}{1}
\caption{TVQA base (no fine-tuning).}
\label{tab:tvqa_base_halfcol}
\begin{tabular}{lr}
\toprule
Setting & Accuracy \\
\midrule
Pooling only & 69.7\% \\
KeyFrame only & 56.8\% \\
\bottomrule
\end{tabular}
\end{table}

Additional qualitative examples are available on the \href{https://povqa.github.io}{project page}, including representative successes and failures on both ReasonVQA and TVQA.
\section{Conclusion}
We presented POVQA for long-video multimodal reasoning under strict context budgets. The method compresses video into 1 fps pooled images, supervises rationale+answer generation with SFT, and optionally applies DPO for preference alignment. In our pilot setting, the main gain comes from pooled evidence plus SFT, while DPO is helpful in some regimes and neutral or negative in others.

ReasonVQA is a pilot diagnostic dataset (12 movies, 239 QA+rationale triplets) intended for controlled analysis rather than large-scale benchmarking. In this setting, we observe strong improvements over pooled-only baselines on answer and rationale metrics, non-trivial zero-shot transfer to TVQA, and interpretable failure patterns tied to temporal compression. Overall, the study emphasizes test-time efficiency and transparent shortcomings analysis in multimodal reasoning.

\noindent\textbf{Limitations.} This study has several limitations: (i) small dataset scale and movie-domain bias, (ii) pooling can lose fine-grained temporal order and tiny action cues, (iii) DPO instability with limited preference pairs, and (iv) metric mismatch between lexical overlap and human-judged reasoning quality.

\noindent\textbf{Future work.} Future directions include expanding the dataset, evaluating on standard long-video benchmarks with stronger video-native baselines, adding judge-based or task-specific accuracy metrics, and testing the same compression recipe on smaller and larger backbones.

{
    \small
    \bibliographystyle{ieeenat_fullname}
    \bibliography{main}
}

\end{document}